\documentclass{esannV2}
\usepackage[dvips]{graphicx}
\usepackage[latin1]{inputenc}
\usepackage{amssymb,amsmath,array}
\usepackage[english]{babel}
\newcommand{\bX}{\mathbf{X}}
\newcommand{\bZ}{\mathbf{Z}}

\DeclareMathOperator{\balpha}{\boldsymbol{\alpha}}
\DeclareMathOperator{\bPi}{\boldsymbol{\Pi}}

\begin{document}
\title{Graphs in machine learning: an introduction}

\author{Pierre Latouche and Fabrice Rossi
%
%
\vspace{.3cm}\\
%
Université Paris 1 Panthéon-Sorbonne - Laboratoire SAMM EA 4543\\
90 rue de Tolbiac, F-75634 Paris Cedex 13 - France
}

\maketitle

\begin{abstract}
Graphs are commonly used to characterise interactions between objects of interest. Because they are based on a straightforward formalism, they are used in many scientific fields from computer science to historical sciences. 
In this paper, we give an introduction to some methods relying on graphs for learning. This includes both unsupervised and supervised methods. Unsupervised learning algorithms usually aim at visualising graphs in latent spaces and/or clustering the nodes. Both focus on extracting knowledge from graph topologies. While most existing techniques are only applicable to static graphs, where edges do not evolve through time, recent developments have shown that they could be extended to deal with evolving networks. 
In a supervised context, one generally aims at inferring labels or numerical values attached to nodes using both the graph and, when they are available, node characteristics. Balancing the two sources of information can be challenging, especially as they can disagree locally or globally. 
In both contexts, supervised and unsupervised, data can be relational (augmented with one or several global graphs) as described above, or graph valued. In this latter case, each object of interest is given as a full graph (possibly completed by other characteristics). In this context, natural tasks include graph clustering (as in producing clusters of graphs rather than clusters of nodes in a single graph), graph classification, etc. 
\end{abstract}

\section{Real networks}

One of the first practical studies on graphs can be dated back to the original
work of Moreno \cite{Moreno} in the 30s.  Since then, there has been a growing interest in
graph analysis  associated with  strong developments in  the modelling
and the processing of these data. Graphs are now used in many scientific fields. In
Biology \cite{Newman2003GraphSurveySIAM,Article:Albert:2002,articlebarabasi2004}, for instance, metabolic networks can describe pathways of
biochemical reactions \cite{articlelacroix2006}, while in social sciences networks are used to
represent         relation         ties         between         actors
\cite{snijders1997estimation,nowicki2001estimation,hoff2002latent,handcock2007model}. Other
examples  include  powergrids   \cite{articlewatts1998}  and  the  web
\cite{articlezanghi2008}.   Recently,   networks   have   also   been
considered  in other  areas such  as geography  \cite{ducruet2013} and
history \cite{rossivillavialanaixetal2014digital-medievalist,RSM}. In machine learning, networks
are  seen as  powerful tools  to model  problems in  order to  extract
information from data and for  prediction purposes. This is the object
of this paper. For more complete surveys, we refer to \cite{goldenberg2010survey,salter2012,matias2014,inbooklatouche2014}.

In this section, we
introduce  notations  and highlight  properties  shared  by most  real
networks.  In  Section  \ref{sec:graphclustering}, we  then  consider
methods aiming  at extracting  information from  a unique  network. We
will particularly  focus on  clustering methods where  the goal  is to
find     clusters     of     vertices.     Finally,     in     Section
\ref{sec:comparegraphs},  techniques that  take a  series of  networks
into  account,  where   each  network  is  seen  as   an  object,  are
investigated. In  particular, distances and kernels for graphs are discussed.

\subsection{Notations}

A graph is first characterised by  a set $\mathcal{V}$ of $N$ vertices
and a set $\mathcal{E}$ of edges  between pairs of vertices. The graph
is said  to be  directed if  the pairs $(i,  j)$ in  $\mathcal{E}$ are
ordered,  undirected otherwise.  A graph  with self  loops is  made of
vertices which can be connected to themselves.  The
degree of a vertex $i$ is the  total number of edges connected to $i$,
with self loops counted twice. In most applications, only the presence
or absence of an edge is characterised. However, edges can also be weighted by a
function    $h:\mathcal{E}    \to     \mathbb{F}$    for    any    set
$\mathbb{F}$. More generally arbitrary labelling functions can be defined on
both the vertices and the edges, leading to labelled graphs. 

A graph is usually described by an $N \times N$ adjacency matrix
$(\bX)_{ij}$  where  $X_{ij}$ is  the  value  associated to  the  edge
between the  $(i, j)$  pair. It  is equal  to zero  in the  absence of
relationship between  the nodes. In  the case of binary  graphs, the
matrix $\bX$ is binary and $X_{ij}=1$ indicates that the two vertices are
connected. If  the graph is directed  then $\bX$ is symmetric  that is
$X_{ij}=X_{ji}$ for all $(i, j)$.

We use interchangeably the vocabulary from graph theory introduced
above and a less formal vocabulary in with a graph is called a network and a
vertex a node. In general, the network is the real world object while the
graph is its mathematical representation, but we have a more relaxed use of
the terms. 

\subsection{Properties}

A remarkable characteristic of most real networks is that they share common
properties
\cite{articledorogovtsev2000,proceedingsamaral2000,articlestrogatz2001,Newman2003GraphSurveySIAM}.
First, most of them are sparse \emph{i.e.} the number of edges present in not
quadratic in the number of vertices, but linear. Thus, the mean degree remains
bounded when $N$ increases and the network density, defined as the ratio
between the number of existing edges over the number of potential edges, tends
to zero.  Second, while some vertices of a real network can have few
connections or no connection at all with the other vertices, most vertices
belong to a single component, so called \emph{giant component}, where it is
always possible to find a path, \emph{i.e.} a set of adjacent connected edges,
connecting any pair of nodes.  Nodes can be disconnected from this component,
forming significantly smaller components.  Finally, we would like to highlight
the degree heterogeneity and small world properties. The first property states
that few vertices have a lot of links, while most of the vertices have few
connections. Therefore, scale free distributions are often considered to model
the degrees \cite{articlebarabasi1999,clauset09power}.  The second one indicates that the
shortest path from one vertex to another is generally rather small, typically
of size $O(\log(N))$.

\section{Graph clustering}\label{sec:graphclustering}

In order to extract information  from a unique graph, unsupervised
methods  usually  look  for  cluster of  vertices  sharing  similar
connection profiles, a particular case of general vertices clustering
\cite{Schaeffer:COSREV2007}. They  differ in the way they  define the topology 
on top of which clusters are built. 

\subsection{Community structure}

Most graph clustering algorithms  aim at uncovering specific types of clusters,
so called communities, where there  are more edges between vertices of
the   same    community   than    between   vertices    of   different
communities. Thus, communities appear in the form of densely connected
clusters of  vertices, with  sparser connections between  groups. They
are  characterised by  the \emph{friend  of  my friend  is my  friend}
effect, \emph{i.e.}  a transitivity property, also  called assortative
mixing effect. Two  families of methods for  community discovering can be
singled out among a vast set of methods \cite{FortunatoSurveyGraphs2010},  depending  on  wether they  maximize  a  score
derived from the modularity score of Girvan and Newman \cite{girvan2002community} or rely on the latent
position  cluster  model  (LPCM)  of  Handcock,  Raftery  and  Tantrum
\cite{handcock2007model}.

\subsubsection{Modularity score}

A series of community detection algorithms have been proposed (see for
instance \cite{girvan2002community,articlenewman2004,articlenewman2004a} and
the survey \cite{FortunatoSurveyGraphs2010}). They involve iterative removal
of edges from the network to detect communities where candidate edges for
removal are chosen according to betweenness measures. All measures rely on the
same idea that two communities, by definition, are joined by a few edges and
therefore, all paths from vertices in one community to vertices in the other
are likely to path along these few edges.  Therefore, the number of paths that
go along an edge is expected to be larger for inter community edges. For
instance, the edge betweenness of an edge does account for the number of
shortest paths between all pairs of vertices that run along that edge.
Moreover, the random walk betweenness evaluates the expected number of times a
random walk would path along the edge, for all pairs of vertices.

The iterative  removal of  edges produces  a dendrogram,
describing  a  hierarchical structure,  from  a  situation where  each
vertex belongs  to a different  cluster to the inverse  scenario where
all vertices are  clustered within the same  community. The modularity
score  \cite{girvan2002community}  is  then  considered  to  select  a
particular  division  of  the  network into  $K$  clusters.   Denoting
$e_{kl}$ the fraction  of edges in the network  connecting vertices of
communities $k$ and $l$, as well as $a_k =\sum_{l=1}^{K}a_{kl}$, the fraction
of edges that  connect with vertices of community  $k$, the modularity
score is given by:
\begin{equation*}
  K_{mod} = \sum_{k=1}^{K}(e_{kk} - a_{k}^2).
\end{equation*}
Such a criterion is computed for the different levels in the hierarchy
and $K$ is chosen such that $K_{mod}$ is maximised.

Rather that  building the  complete dendrogram, other  algorithms have
focused  on optimising  the  modularity score  directly, as it is beneficial
both in computational terms and in the perceived quality of the obtained
partitions. A very popular algorithm, the so-called Louvain method
\cite{BlondelEtAll2008}, proceeds by a series of greedy exchanges and merging
that turns a fully refined partition into a coarser one that provides a
(local) maximum of the modularity. Better solutions can be obtained using more
sophisticated heuristics \cite{NoackRotta2009MultiLevelModularity} but
maximising the modularity is a NP-hard problem \cite{brandes2008modularity}. 

Note that modularity
approaches   have    been   shown   to   be    asymptotically   biased
\cite{bickel2009nonparametric}. To tackle this issue, degree corrected
methods were  introduced in order  to take  the degrees of  nodes into
account.

\subsubsection{Latent position cluster model}

Alternative  approaches,   looking  for  clusters  of   vertices  with
assortative mixing, usually rely on the LPCM
model \cite{handcock2007model} which is a generalisation of the latent
position model (LPM) \cite{hoff2002latent}. In the original LPM model,
each vertex $i$ is first assumed  to be associated  with a position $\bZ_{i}$ in a
Euclidean latent space $\mathbb{R}^{d}$. Each edge between a pair $(i, j)$ of vertices
is  then drawn  depending  on $\bZ_{i}$  and  $\bZ_{j}$. Both  maximum
likelihood  and  Markov  chain  Monte  Carlo  (MCMC)  techniques  were
considered  to   estimate  the  model  parameters   and  the  latent
positions.  The  corresponding  mapping   of  the  vertices  into  the
Euclidean latent space  produces a representation of  the network such
that   nodes  which   are  likely   to  be   connected  have   similar
positions. Note that if the latent space is low dimensional, typically
of  dimension $d=1,2,3$,  then  the representation  can be  visualised
which is feature appreciated by practitioners.

The LPM model was extended in  order to look for both a representation
of  the   network  and  a   clustering  of  the  vertices.   Thus,  the
corresponding LPCM model  assumes that the positions are  drawn from a
Gaussian mixture  model in  the latent space  such that  each Gaussian
distribution corresponds to a cluster.  A two stage maximum likelihood
approach along with a Bayesian MCMC scheme were proposed for inference
purposes.  Moreover,  conditional  Bayes factors  were  considered  to
estimate the number of clusters from the data. Finally variational bayesian
inference is also possible \cite{salter2013variational}.

\subsection{Heterogeneous structures}

So far, we  have discussed methods looking  exclusively for communities
in networks. Other approaches usually derive from the stochastic block
model (SBM) of Nowicki and Snijders \cite{nowicki2001estimation}. They
can also look for communities, but not only. 

The SBM models  assumes that nodes are spread in  unknown clusters and
that the  probability of a  connection between  two nodes $i$  and $j$
depends on their  corresponding clusters. In practice,  a latent vector
$\bZ_{i}$  is drawn  from a  multinomial distribution  with parameters
$(1, \balpha=\{\alpha_{1},\dots,\alpha_{K}\})$, where $\alpha_{k}$ is the proportion of cluster $k$. Therefore,
$\bZ_{i}$ is a binary vector of size  $K$ with a single $1$, such that
$Z_{ik}=1$   indicates  that   $i$   belongs  to   cluster  $k$,   $0$
otherwise. If $i$ is in cluster $k$ and $j$ in $l$, then the SBM model
assumes that there is a probability $\pi_{kl}$ of a connection between
the two nodes. All connection probabilities are characterised by a $K
\times  K$ matrix  $\bPi$.  Note  that a  community  structure can  be
defined by setting values for the diagonal terms 
of $\bPi$ to higher values than extra diagonal terms \cite{hofman2008bayesian}.  In
practice, because
no  assumptions are  made regarding  $\bPi$,  the SBM  model can  take
heterogeneous structures into account \cite{daudin2008mixture,inbooklatouche2009,articlelatouche2012}.

While generating a network with such a sampling scheme
is  straightforward, estimating  the  model  parameters $\balpha$  and
$\bPi$  as  well  as  the  set $(\bZ)_i$  of  all  latent  vectors  is
challenging. One of the key issue is that the posterior distribution
of $\bZ$  given the  adjacency matrix $\bX$  and the  model parameters
$(\balpha,   \bPi)$   cannot   be  factorised   due   to   conditional
dependency. Therefore,  standard optimisation  algorithms, such  as the
expectation maximisation (EM) algorithm,  cannot be derived. To tackle
this  issue  variational  and   stochastic  approximations  have  been
proposed. Thus,  \cite{daudin2008mixture} relied  on a  variational EM
(VEM) algorithm whereas  \cite{articlelatouche2012} used a variational
Bayes EM (VBEM)  approach. Alternatively, \cite{nowicki2001estimation}
estimated  the  posterior distribution  of  the  model parameters  and
$\bZ$, given $\bX$,
by considering Gibbs sampling. 

A even more fondamental question concerns the estimation of the number
of clusters present  in the data. Unfortunately,  since the likelihood
is not tractable either, standard model selection criteria, like the
Akaike information criterion (AIC) or  the Bayesian IC (BIC) cannot be
computed.  Again, variational along with asymptotic Laplace approximations were derived to obtain
approximate            model             selection            criteria
\cite{daudin2008mixture,articlelatouche2012}.

In some cases,  the clustering of the nodes and  the estimation of the
number of  clusters are performed  at the same time  using allocation
sampler \cite{mcdaid11}, greedy  search \cite{articlecome2013}, or non
parametric schemes \cite{kemp2006learning}.

\subsection{Extensions}

Since the original development of  the SBM model, many extensions have
been   proposed    to   deal   for   instance    with   valued   edges
\cite{articlemariadassou2010}  or  to   take  into  account  covariate
information  \cite{articlezanghi2010,matias2014}. The  random subgraph
model (RSM) \cite{RSM} for instance assumes that a partition of the nodes into
subgraphs is observed and that the subgraphs are made of (unknown) latent
clusters, as  in the SBM  model, with various mixing  proportions. The
edges are typed. In
parallel, strategies looking for overlapping clusters, where each node
can   belong   to   multiple   clusters,   have   been   derived.   In
\cite{airoldi2008mixed}, a vertex $i$ belongs a cluster in its relation
with a given vertex $j$. Because $i$ is involved in multiple relations
in  the  network,  it  can  belong   to  more  than  one  cluster.  In
\cite{latouche2009overlapping},  the multinomial  distribution of  the
SBM  model  is replaced  with  a  product of  Bernoulli  distribution,
allowing each vertex to belong to no, one, or several clusters. 

In the last few years, a lot of attention has been paid on extending the
approaches mentioned previously in order to deal with dynamic networks where
nodes and/or edges can evolve through time.  The main idea consists in
introducing temporal processes, such as hidden Markov model (HMM) or linear
dynamic systems \cite{xing2010state,yang2011detecting,xu2013dynamic}.  While
models usually focus on modelling the dynamic of networks through the
evolution of their latent structures, Heaukulani and Gharamani
\cite{heaukulani2013dynamic} chose to define how observed social interactions
can affect future unobserved latent structures.  We would also like to highlight
the work of Dubois, Butts, and P. Smyth \cite{proceedingsdubois2013}.
Contrary to most dynamic clustering approaches, they considered a non
homogeneous Poisson process allowing to deal with a continuous time periods
where events, \emph{i.e.} the creation or removal of an edge, can occur one at
a time. Another approach for graph clustering in the continuous time context
is provided by \cite{guigouresboulleetal2012triclustering-approach} which
builds a coclustering structure on the vertices of the graph and on the time
stamps of the edges.


\section{Multiple graphs}\label{sec:comparegraphs}
While a large part of the graph related literature in machine learning targets
the case of a single graph, numerous applications lead naturally to data sets
made of graphs, that is situations in which each data point is a graph (or
consists in several components including at least one graph). This is the case
for instance in chemistry where molecules can be represented by undirected
labelled graphs (see e.g. \cite{Ralaivola20051093}) and in biology where the
structure of a protein can be represented by a graph that encodes
neighborhoods between it fragments as in \cite{borgwardt2005protein}. In fact,
the use of graphs as structured representations of complex data follows a long
tradition with early examples appearing in the late seventies
\cite{RouvaryBalaban1979} and with a tendency to become pervasive in the last
decade. 

It should be noted that even in the case of a single global graph
described in the first part of this paper, it is quite natural to study
multiple graphs derived from the global one, in particular via the
ego-centered approach which is very common in social sciences (see
e.g. \cite{JCC4:JCC401}). The main idea is to extract from a large social
network a set of small networks centered on each of the vertices under
study. For real world social networks, it is in general the only possible
course of action, the whole network being impossible to observe (see
e.g. \cite{dhanjalblanchemancheetal2012dissemination-health} for an example).  

When dealing with multiple graphs, one tackles the traditional tasks of
machine learning, from unsupervised problems (clustering, frequent patterns
analysis, etc.)
to supervised ones (classification, regression, etc.). There are two main tendencies
in the literature: the design of specialized methods obtained by adapting
classical ones to graphs and the use of distances and kernels coupled with
generic methods.

\subsection{Specialized methods}
As graphs are not vector data, classical machine learning techniques do not
apply directly. Numerous methods have been adapted in rather specific ways to
handle graphs and other non vector data, especially in the neural network
community \cite{HammerJain04NonStandardData,cottrellolteanuetal2012neural-networks}, for instance via recursive
neural networks as in
\cite{HammerEtAl2004StructuredNeurocomputing,Hagenbuchner20091419}. In those
approaches, each graph is processed vertex by vertex, by leveraging the
structure to build a form of abstract time. The recursive model maintains an
implicit knowledge of the vertices already processed by means of its space
state neurons. 

\subsection{Distances and kernels}
A somewhat more generic solution consists in building distances (or
dissimilarities) between graphs and then in using distances based methods
(such as methods based on the so-called relational approach
\cite{hammerhasenfussetal2007topographic-processing}). One difficulty is that
graph isomorphism should be accounted for when two graphs are compared: two
graphs are isomorphic if they are equal up to a relabeling of their
vertices. Any sound dissimilarity/distance between graphs should detect
isomorphic graphs. This is however far more complex than expected
\cite{fortin1996graph} up to a point that the actual complexity class of the
graph isomorphism problem remains unknown (it belongs to the NP class but not
to the NP-complete class, for instance). While exact algorithms appear fast on
real world graphs, their worst case complexities are exponential, with a best bound in
$O(2^{\sqrt{n\log n}})$ \cite{Luks198242}. In addition, subgraph isomorphism,
i.e. determining whether a given graph contains a subgraph that is isomorphic
to another graph is NP-complete.

Nevertheless, numerous exact or approximate algorithms have been defined to
try and solve the (sub)graph isomorphism problem (see
e.g. \cite{bunke2000graph}). In particular, it has been shown that those
problems (and related ones) are special cases of the computation of the graph
edit distance \cite{790431}, a generalization of the string edit distance
\cite{Levenhstein}. The graph edit distance is defined by first introducing
edition operations on graph, such as insertion, deletion and substitution of
edges and vertices (labels and weights included). Each operation is assigned a
numeric cost. The total cost of a series of operations is simply the sum of
the individual costs. Then the graph edit distance between two graphs is the
cost of the least costly sequence of operations that transforms one of the
graph into the other one (see \cite{gao2010survey} for a survey). 

A rather different line of research has provided a set of tools to compare
graphs by means of kernels (as in reproducing kernels
\cite{Aronszajn1950}). Those symmetric positive definite similarity functions
allow one to generalize any classical vector space method to non vector data
in a straightforward way \cite{ScholkopfSmola2002Book}. Numerous of such
kernels have been defined to compare two graphs
\cite{vishwanathan2010graph}. Most of them are based on random walks that take
place on the product of the two graphs under comparison. 

Once a kernel or a distance has been chosen, one can apply any of the kernel
methods \cite{ShaweTaylorChristianini2004KernelMethods} or of the relational
methods \cite{hammerhasenfussetal2007topographic-processing}, which gives
access to support vector machine, kernel ridge regression and kernel k-means
to cite only a few. 

\section{Conclusion}
This paper has only scraped the surface of the vast literature about graphs in
machine learning. Complete area of graph applications in machine learning were
ignored. 

For instance, it is well know that extracting a neighborhood graph from a
classical vector data set is an efficient way to get insights on the topology
of the data set. This has led to numerous interesting applications ranging
from visualization (as in isomap \cite{IsomapScience2000} and its successors)
to semi-supervised learning \cite{belkin2006manifold}, going through spectral
clustering \cite{von2007tutorial} and exploratory analysis of labelled data
sets \cite{aupetit2005high}.

Another interesting area concerns the so-called relational data framework when
a classical data set is augmented with a graph structure: the vertices of the
graph are elements of a standard vector space and are thus traditional data
points, but they are interconnected via a graph structure (or several ones in
complex settings). The challenge consists here in taking into account the
graph structure while processing the classical data or vice-versa in taking into
account the data point descriptions when processing the graph. Among other
issues, those two different sources of information can be contradictory for a
given task. A typical application of such a framework consists in annotating
nodes on social media \cite{jacob2011classification}.

While we have presented some temporal extensions of classical graph related
problems, we have ignored most of them. For instance, the issue of information
propagation on graphs has received a lot of attention
\cite{rodriguez2014uncovering}. Among other tasks, machine learning can be used e.g. to predict the
probability of passing information from one actor to another as in
\cite{dhanjalblanchemancheetal2012dissemination-health}. 

More generally, the massive spread in the last decade of online social
networking has the obvious consequence of generating very large relational data
sets. While non vector data have been studied for quite a long time, those new
data sets push the complexity one step further by mixing several types of non
vector data. Objects under study are now described by complex mixed data
(texts, images, etc.) and are related by several networks (friendship, online
discussion, etc.). In addition, the temporal dynamic of those data cannot be
easily ignored or summarized. It seems therefore that the next set of problems
faced by the machine learning community will include graphs in numerous forms,
including dynamic ones. 

\begin{footnotesize}
\bibliographystyle{abbrv}
\bibliography{bib}

\end{footnotesize}

\end{document}